\documentclass{article}
\usepackage[table]{xcolor}

\usepackage{spconf,amsmath,graphicx}
\usepackage{times}
\usepackage{graphicx}
\usepackage{amssymb}
\usepackage{url,hyperref}
\usepackage{enumitem}
\usepackage[utf8]{inputenc} 
\usepackage[T1]{fontenc}    
\usepackage{hyperref}       
\usepackage{url}            
\usepackage{booktabs}       
\usepackage{amsfonts}       
\usepackage{nicefrac}       
\usepackage{microtype}      
\usepackage{lipsum}
\usepackage{cite}
\usepackage{amsmath,amssymb,amsfonts}
\usepackage{algorithmic}
\usepackage{graphicx}
\usepackage{graphics} 
\usepackage{multirow}
\usepackage{cite}
\usepackage{verbatim}
\usepackage{url}
\usepackage{ctable}
\usepackage[normalem]{ulem}
\usepackage{amsmath,amssymb,amsfonts}
\usepackage{algorithmic}
\usepackage{graphicx}
\usepackage{textcomp}
\usepackage{xcolor}
\usepackage{tabularx}
\usepackage{makecell}
\usepackage{mathtools}
\usepackage{commath}
\usepackage{array,multirow}
\usepackage{textcomp}
\usepackage{xcolor}
\usepackage{import}
\usepackage[normalem]{ulem}

\usepackage{cite}
\usepackage{amsmath,amssymb,amsfonts}
\usepackage{algorithmic}
\usepackage{graphicx}
\usepackage{textcomp}
\usepackage{xcolor}
\usepackage{tabu}
\usepackage{graphicx}
\usepackage{amsmath}

\usepackage[table]{xcolor}

\title{Self-supervised Learning for ECG-based Emotion Recognition}

\name{Pritam Sarkar, Ali Etemad}
\address{Department of Electrical and Computer Engineering\\Queen's University, Kingston, Ontario, Canada\\ \small{\texttt{\{pritam.sarkar, ali.etemad\}@queensu.ca}}}

\begin{document}

\maketitle

\begin{abstract}
We present an electrocardiogram (ECG) -based emotion recognition system using self-supervised learning. Our proposed architecture consists of two main networks, a signal transformation recognition network and an emotion recognition network. First, unlabelled data are used to successfully train the former network to detect specific pre-determined signal transformations in the self-supervised learning step. Next, the weights of the convolutional layers of this network are transferred to the emotion recognition network, and two dense layers are trained in order to classify arousal and valence scores. We show that our self-supervised approach helps the model learn the ECG feature manifold required for emotion recognition, performing equal or better than the fully-supervised version of the model. Our proposed method outperforms the state-of-the-art in ECG-based emotion recognition with two publicly available datasets, SWELL and AMIGOS. Further analysis highlights the advantage of our self-supervised approach in requiring significantly less data to achieve acceptable results.

\end{abstract}

\begin{keywords}
Self-supervised Learning, Multi-task, Emotion Recognition, ECG
\end{keywords}

\section{Introduction}
Electrocardiogram (ECG) has been proven to be a reliable source of information for emotion recognition systems \cite{picard2001toward, nardelli2015recognizing, pritam_acii, nussinovitch2011reliability}. Automated ECG analysis can identify the affective states of users such as happiness, sadness, and stress, among others. Understanding and quantifying the emotional states of humans can have significant effects on intelligent human-machine systems. ECG and other physiological signals have been used in several affective computing applications. For example, \cite{healey2005detecting} performed stress detection using ECG, electromyography (EMG), and galvanic skin response (GSR), during driving tasks. In \cite{liu2009dynamic}, a dynamic difficulty adjustment mechanism for computer games was proposed to provide tailored gaming experience to individual users by analysing ECG and GSR. An ECG-based deep multitask learning framework was proposed in \cite{pritam_acii, sensor_kyle} for adaptive simulation. The aim was to provide personalised training experience to individual users based on their level of expertise and cognitive load. 

Although ECG has considerable potential for affective computing, we often lack sufficient labelled data in order to train deep supervised models. To tackle this problem, we propose a deep learning solution based on self-supervised learning \cite{raina2007self}. Self-supervised learning is a representation learning approach, in which models are trained using automatically generated labels instead of human annotated labels. There are a number of advantages to self-supervised learning. First, the feature manifolds learnt using this approach are often invariant to inter-instance and intra-instance variations \cite{wang2017transitive} by learning more generalized features rather than task-specific ones. As a result, these models can be reused for different tasks within the same domain. Moreover, self-supervised models require less amount of human-annotated labels to achieve high classification performance.

In this paper we propose ECG-based emotion recognition using multi-task self-supervised learning for the first time. We use two publicly available datasets, SWELL \cite{swell_dataset} and AMIGOS \cite{amigos_dataset}. First, to train our network with automatically generated labels, we perform $6$ different signal transformation tasks. Next, the $6$ transformed signals along with the original ones are used to train a multi-task convolutional neural network (CNN) in a self-supervised manner. The proposed CNN architecture consists of $3$ convolutional blocks as shared layers followed by $2$ task-specific dense layers. As the next step, we use the pre-trained model for emotion classification. To do so, we transfer the weights of the pre-trained network to a new network and train a simple fully-connected layer, and test the framework on the two datasets. Our analysis shows that our self-supervised model is better or competitive in comparison to the same network when trained in fully supervised fashion. Finally, we set a new state-of-the-art for arousal and valence detection for SWELL and AMIGOS datasets.

\section{Related Work}
Self-supervised learning is becoming popular in the field of computer vision \cite{wang2017transitive, kocabas2019self}, natural language processing \cite{wu2010open}, speech and signal processing \cite{tagliasacchi2019self, multitask_selfsupervised}, and others. Human activity recognition using self-supervised learning was performed in \cite{multitask_selfsupervised}, where different signal transformation tasks were carried out as pretext tasks to generate automatic labels. A wide variety of activity recognition tasks such as walking, sitting, and jogging were performed using $6$ different publicly available datasets. It was shown in \cite{multitask_selfsupervised} that self-supervised learning helps convolution networks learn high-level features. Another work \cite{kocabas2019self} performed $3$D pose estimation using self-supervised learning. Training a model to estimate $3$D poses requires large amounts of training data and is highly resource-dependent. To overcome these problems, the authors used available $2$D pose data and performed epipolar geometry to calculate $3$D poses in self-supervised manner. The obtained $3$D poses were used to train a model to perform $3$D pose estimation. In \cite{Xu_2019_CVPR}, a self-supervised learning method was used for action recognition. The model was first trained using a $3$D convolution neural network to predict the order of shuffled video clips. Subsequently, the pre-trained model was fine-tuned using nearest neighbour technique for action recognition.

\section{Proposed Method}
\subsection{Self-supervised Learning}
\begin{figure}[t]
    \centering
    \includegraphics[width=0.7\columnwidth]{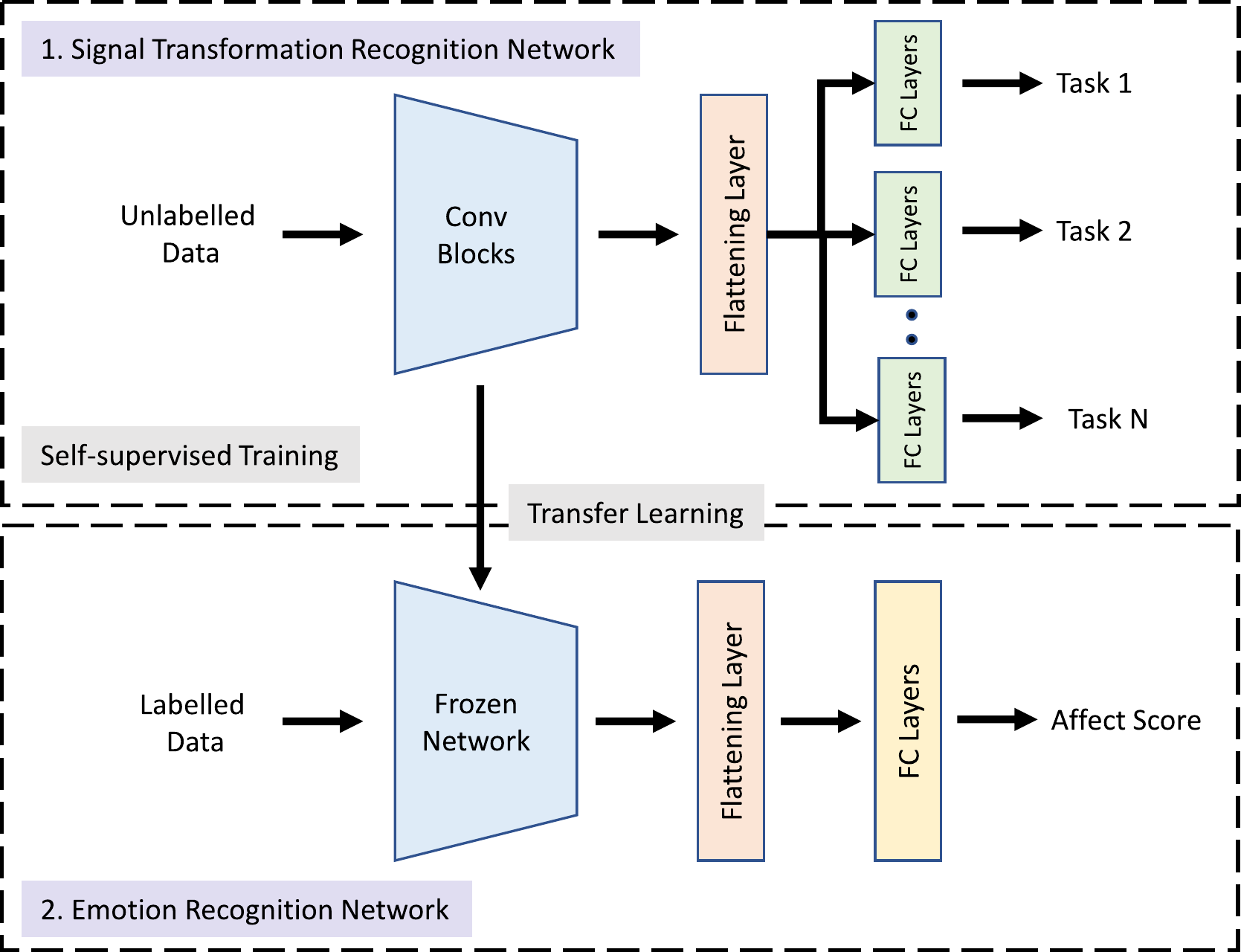}
    \caption{The proposed self-supervised architecture is presented. First, a multi-task CNN is trained using automatically generated labels to recognize signal transformations. Then, the weights are transferred to the emotion recognition network, where fully connected layers are trained to classify emotions.}
    \label{fig:small_ssl}
\end{figure}

Let $T_{p}$ and $T_{d}$ be two categories of tasks, namely \textit{pretext} and \textit{downstream} tasks, where $T_{p}$ is trained using automatically generated pseudo-labels $ P{_j}$ while $T_{d}$ is trained using true labels ${y}_{i}$. Now, let $(X{_j}, P{_j})$ be an example tuple of inputs and pseudo-labels for $T_{p}$, $j\in[0,N]$, where $0$ denotes the original signal and $1, 2, .., N$ correspond to the number of performed signal transformations.
Our goal is to obtain a feature manifold $F$ that can easily distinguish between the $T_{d}$ classes. To do so, we define a model $\psi_j, F = \gamma(X_j, \theta)$, where $\theta$ is the set of trainable parameters and $\psi_j$ is the predicted probability of the $j^{th}$ transformation task. Accordingly, we find the optimum parameters $\theta$ by minimizing the weighted average of the individual losses of the signal transformation network ${L}_{j}$, where  \({L}_{j} = [P_j \log \psi_j + (1 - P_j) \log (1 - \psi_j)]\). The total loss is represented as \(\sum_{j=0}^{N}{ {\alpha}_{j} {L}_{j}}\)
where ${\alpha}_{j}$ is the loss coefficient of the $j^{th}$ task. Consequently, we can use the feature manifold $F$ to perform $T_{d}$, since it contains useful information regarding the original signals $X_{0}$. 
In order to perform $T_{d}$ using the learned feature manifold $F$, we use a simple model $\rho = \zeta(F, \theta')$, where $\theta'$ is the set of trainable parameters, and $\rho$ is the probability vector of $T_{d}$ classes. We then calculate the optimum value of $\theta'$ by minimizing the cross entropy loss, which can be mathematically expressed as \(\sum_{i=1}^{M}{ {y}_{i} \log{\rho_i}}\),
where $M$ is total number of classes. Figure \ref{fig:small_ssl} presents an overview of our pipeline, where self-supervised training is performed followed by transfer learning for emotion recognition.

\subsection{Signal Transformation Tasks}
We train our self-supervised network on the pretext tasks to enable the network to learn spatiotemporal features and abstract representations of the data. For our developed signal transformation recognition network, $6$ different transformations are performed \cite{multitask_selfsupervised, um2017data}. These transformations are as follows:
\begin{itemize}[noitemsep, leftmargin=*]
    \item \textbf{Noise addition:} Random Gaussian noise is added to the ECG signal. 
    \item \textbf{Scaling:} The magnitude of the ECG Signal is scaled by $20\%$.
    \item \textbf{Negation:} The amplitude of the ECG signal is multiplied by $-1$, causing a vertical flip of the original signal.
    \item \textbf{Horizontal flipping:} The ECG signal is flipped horizontally along the time axis.
    \item \textbf{Permutation:} ECG segments are divided into $10$ sub-segments and shuffled, randomly perturbing their temporal locations.
    \item \textbf{Time-warping:} Random segments of ECG signals are stretched and squeezed along the $x$ axis.
\end{itemize}

The above-mentioned transformations are performed on the original ECG signals and stacked to create the input matrix while the corresponding labels of the transformations $0, 1, ..., 6$ are stacked to create the corresponding output vector.The parameters for the signal transformations, for instance the scaling factor and amount of noise, are selected empirically with the goal of maximizing the final emotion recognition performance.

\subsection{Network Architecture}
\textbf{Signal Transformation Recognition Network:}
Our multi-task signal transformation recognition network consists of $3$ convolutional blocks and $2$ dense layers. The convolutional layers are shared for the different tasks while the dense layers are task-specific as shown in Figure \ref{fig:small_ssl}. Each convolutional block consists of $2\times1$D convolution layers with ReLu activation functions followed by a max-pooling layer of size 8. In the convolutional layers, we gradually increase the number of filters, from $32$ to $64$ and $128$. The kernel size is gradually decreased after each convolutional block from $32$ to $16$ and $8$ respectively. Finally, at the end of the convolutional layers, global max-pooling is performed. The dense layers that follow consist of $2$ fully-connected layers with $128$ hidden nodes followed by a sigmoid layer. We use $60\%$ dropout in the dense layers and L$2$ regularization with $\beta = 0.0001$ to overcome possible overfitting. The summary of the network architecture is presented in Table \ref{tab:SSL_arch}.

\textbf{Emotion Recognition Network:} In this step, we develop a simple emotion recognition network with identical convolution layers to the signal transformation recognition network and $2$ dense layers with $64$ hidden nodes, followed by a sigmoid layer. We then transfer the weights from the convolutional layers of the signal transformation recognition network to the convolution layers of this network. The ECG signals and emotion labels are then used as inputs and outputs for training this network. It should be noted that the weights transferred from the convolutional layers of the signal transformation recognition network are frozen and hence not re-trained (only the dense layer is trained). We keep the fully-connected layers of the network simple in order to be able to evaluate the performance of our approach with regards to the self-supervised learning of the signal transformation tasks.

\section{Experiments and Results}
\subsection{Datasets}
We use two public datasets, SWELL \cite{swell_dataset} and AMIGOS \cite{amigos_dataset}. The SWELL dataset was collected from $25$ participants with the aim of understanding the mental stress and emotional attributes of employees in a typical office environment under different working conditions. Each of these conditions were designed for a duration of $30$-$45$ minutes. The AMIGOS dataset was collected from $40$ participants, where emotional video clips were shown to participants individually and in groups to elicit affective reactions. In both the dataset, participants' self-assessed affect scores were recorded on a scale of $1$-$9$ for both arousal and valence. The SWELL dataset has been recorded using MOBI devices (TMSI) \cite{TMSi13:online} with self-adhesive electrodes, at a sampling frequency of $2048$ Hz while the AMIGOS has been collected using Shimmer sensors \cite{shimmer} at a sampling frequency of $256$ Hz. We perform very little pre-processing on the ECG signals. Since the two datasets are recorded at different sampling rates, SWELL ECG signals are first downsampled to $256$ Hz. Then we remove ECG baseline wander for both datasets by applying a high-pass IIR filter with a pass-band frequency of $0.8$ Hz.



\begin{table}[]
    \footnotesize
    \centering
    \caption{The architecture of the signal transformation recognition network is presented.}
    \begin{tabular}{l|l|l}
    \hline
     \textbf{Module} & \textbf{Layer Details} & \textbf{Feature Shape} \\ \hline \hline
      Input & $-$ & $2560 \times 1$ \\ \hline
      \multirow{7}{*}{Shared Layers} & $[\textit{conv}, 1 \times 32, 32] \times 2$ &   $2560 \times 32$  \\ 
                                     &  $[\textit{maxpool}, 1 \times 8, \textit{stride} = 2]$ &   $1277 \times 32$  \\ 
                                     &  $[\textit{conv}, 1 \times 16, 64] \times 2$ &   $1277 \times 64$  \\ 
                                     &  $[\textit{maxpool}, 1 \times 8, \textit{stride} = 2]$ &   $635 \times 64$  \\ 
                                     &  $[\textit{conv}, 1 \times 8, 128] \times 2$ &   $635 \times 128$ \\ \cline{2-3} 
                                     & \textit{global max pooling} &  $1 \times 128$ \\ \hline 
      {\makecell[l]{Task-Specific\\Layers}} & \makecell[l]{$[\textit{dense}] \times 2$\\ $\times$ \textit{$7$ parallel tasks}} & 128  \\ \hline 
      Output & $-$ & $2$ \\ \hline

    \end{tabular}
    \label{tab:SSL_arch}
\end{table}

\subsection{Model Training}
The ECG signals are segmented into a fixed window size of $10$ seconds. Each segment is used to generate 6 transformed variations used to train the signal transformation recognition network. Our proposed architecture is implemented using TensorFlow on an Nvidia 2070 Ti GPU. To train both networks (signal transformation recognition and emotion recognition), Adam optimizer \cite{kingma2014adam} is used with a learning rate of $0.001$ and batch size of $128$. The signal transformation recognition network is trained for $30$ epochs, while the emotion recognition network is trained for $100$ epochs, as steady states are reached with different number of epochs. Similar to \cite{amigos_dataset, santamaria2018using, verma2019comprehensive, sriramprakash2017stress} the affective attributes (output labels) for both datasets are converted to binary classes by setting a threshold equal to the mean. We use $10$-fold cross-validation to evaluate the performance of our proposed model.

\subsection{Performance and Comparison}
\begin{figure}
    \centering
    \includegraphics[width= 0.65\columnwidth]{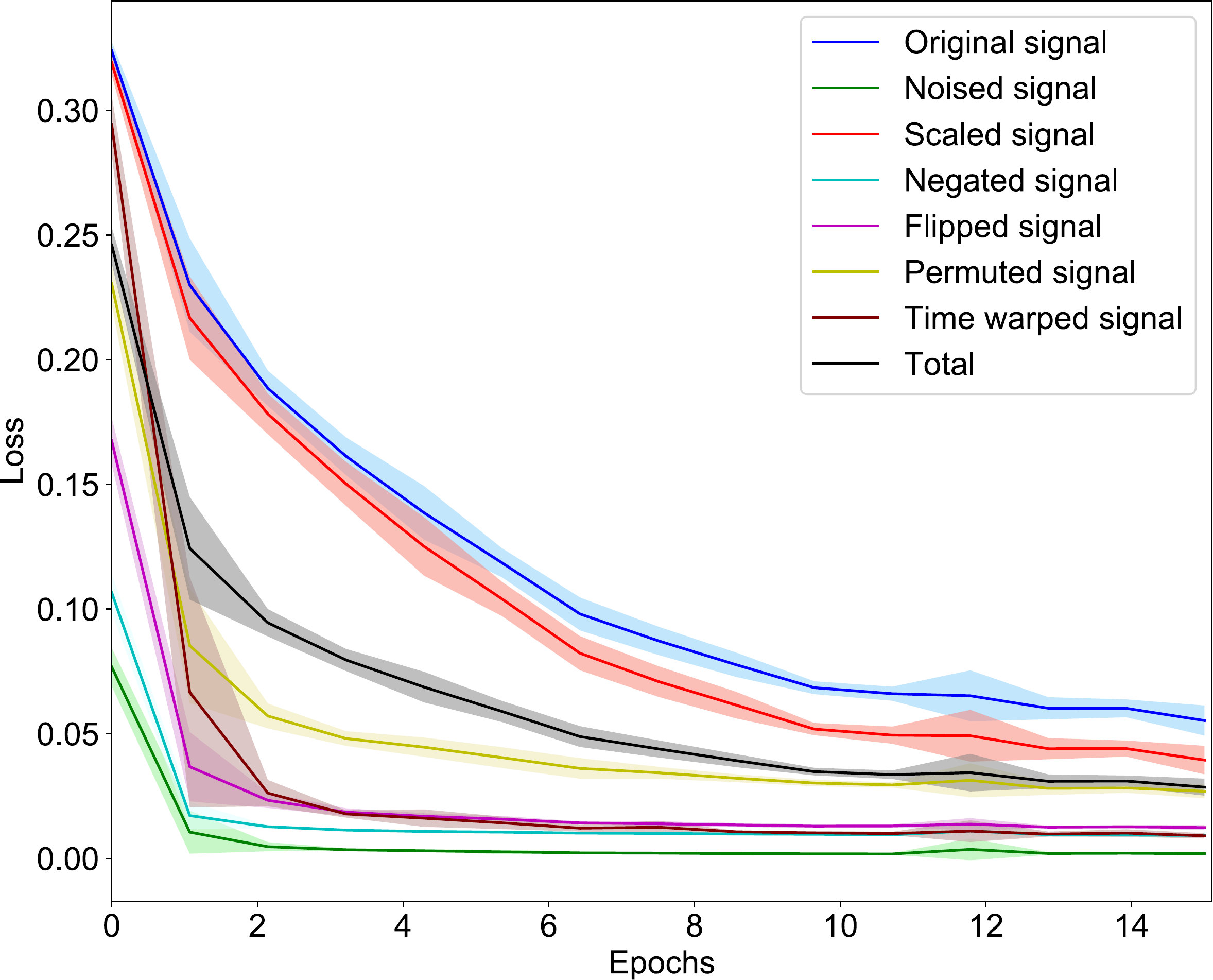}
    \caption{Individual training losses, and total output loss vs. epoch curves are presented for the signal transformation recognition network. The average and standard deviation values are obtained from the $10$ folds of the training phase.}
    \label{fig:training_tf_loss}
\end{figure}

Figure \ref{fig:training_tf_loss} shows that at $13$ epochs, the signal transformation recognition model reaches a steady-state when learning the signal transformation tasks. Clear differences among the steady-state loss of the different transformations are observed, pointing to varied difficulties in the self-supervised training tasks. These tasks allow for learning of ECG-specific representations that, when transferred to the emotion recognition network, aid in classification of emotions with higher accuracy.

Tables \ref{tab:ssl_swell} and \ref{tab:ssl_amigos} show the performance of our self-supervised approach for emotion classification. The results show that for classification of arousal and valence, our model achieves accuracies of $96\%$ and $95.6\%$ with SWELL, while accuracies of $85.1\%$ and $84\%$ are achieved with AMIGOS. To further evaluate the performance of our model, we compare the results with a \textit{fully}-supervised version of the emotion recognition network when trained only using the labeled dataset. See Tables \ref{tab:ssl_swell} and \ref{tab:ssl_amigos}. The comparison shows that the self-supervised approach performs competitive to or better than the fully-supervised method, indicating the effectiveness of our method.

Next, we compare our results with prior work on emotion recognition performed on these two datasets. It should be noted that prior works on the SWELL dataset using the ECG modality have mostly focused on stress detection as opposed to classification of arousal and valence. In \cite{koldijk2016detecting}, this dataset was used to perform binary classification of stress levels using support vector machines (SVM), reporting a baseline accuracy of $64.1\%$ when using ECG and GSR. Similarly in \cite{sriramprakash2017stress}, an SVM classifier was used to perform stress detection, reporting an accuracy of $86.36\%$. A similar task was performed in \cite{verma2019comprehensive} using Bayesian belief network (BBN), reporting an accuracy of $92.6\%$. While estimation of arousal and valence has been performed in \cite{alberdi2018using}, the problem was formulated as regression, thus preventing a valid comparison with our classification approach. As a result, we also performed stress detection on this dataset and achieved an accuracy of $98.3\%$ and $98.4\%$ in self-supervised and fully-supervised methods respectively. Table \ref{tab:ssl_swell} compares our self-supervised model and past supervised works on the same dataset, showing that the proposed model performs with higher accuracy.

For the AMIGOS dataset, baseline classification results were provided in \cite{amigos_dataset}, where classification of arousal and valence was carried out using a Gaussian Naive Bayes classifier, reporting F1 scores of $54.5\%$ and $55.1\%$ for the two tasks respectively. In \cite{santamaria2018using}, a CNN was used to perform classification, reporting accuracies of $81\%$ and $71\%$ for arousal and valence respectively. Other works such as \cite{harper2019bayesian} have also performed emotion recognition on AMIGOS. However, different validation schemes are utilized. Table \ref{tab:ssl_amigos} presents our results in comparison to prior work, once again showing that the self-supervised approach outperforms prior work on the same dataset.

\begin{figure}[t!]
    \centering
    \includegraphics[width=0.75\columnwidth]{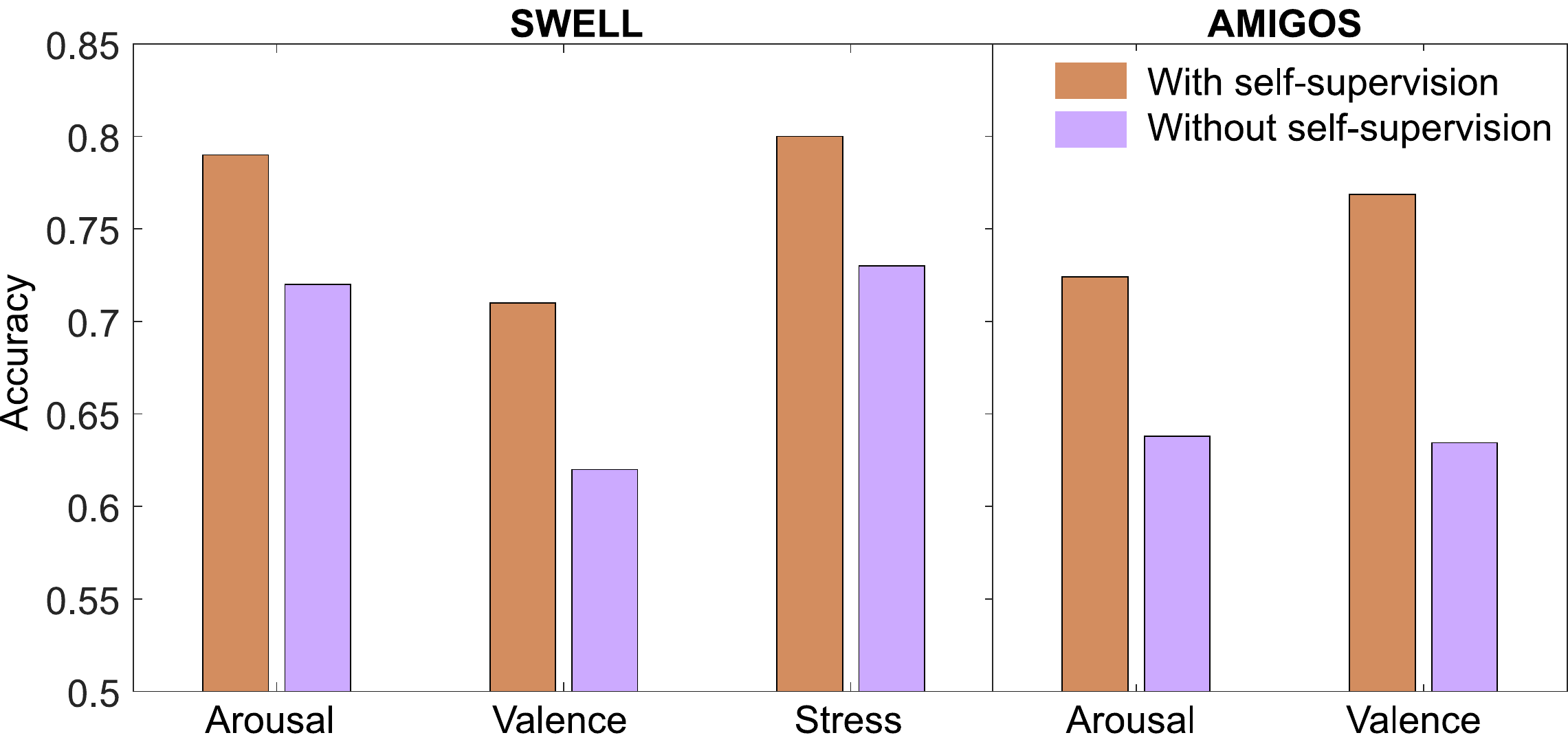}
    \caption{Performance of our method with and without the self-supervised learning step using $1\%$ of the labels in the datasets are presented.}
    \label{fig:less_data}
\end{figure}

Next, in order to evaluate the impact of the self-supervised method on the amount of \textit{labeled data} required to adequately train a model, we utilize only $1\%$ of the labeled data for training both the self-supervised and fully-supervised classification methods. We first use the entire \textit{unlabeled} dataset to train the signal transformation recognition network using the automatically generated labels (signal transformations). Then, $1\%$ of the \textit{labeled data} per-user per-class is used to train and test the emotion recognition network after transferring the weights from the self-supervised network. Next, we also used the same $1\%$-dataset to train a separate CNN without the self-supervised step. Figure \ref{fig:less_data} shows the results where the fully-supervised network trained using the very small dataset often performs considerably worse than our self-supervised model.

\renewcommand{\tabcolsep}{5pt}

\begin{table}[!t]
    \centering
    \footnotesize
    \caption{The results of our self-supervised method on the SWELL dataset are presented and compared to prior work as well as the emotion recognition network without the self-supervised step.}
    \begin{tabular}{@{}l|l|l|l|l|l|l}
        \hline
            \multirow{2}{*}{\textbf{Ref.}} & \multirow{2}{*}{\textbf{Method}} & \makecell{\multirow{2}{*}{\textbf{Stress}}} & \multicolumn{2}{l|}{\textbf{Arousal}} & \multicolumn{2}{l}{\textbf{Valence}} \\ \cline{4-7}
             &&& \textbf{Acc.} & \textbf{F1} & \textbf{Acc.} & \textbf{F1} \\ \hline\hline
            \cite{koldijk2016detecting} & SVM & $0.641$ & \cellcolor{gray!25} & \cellcolor{gray!25} & \cellcolor{gray!25} & \cellcolor{gray!25} \\ \hline  
            \cite{sriramprakash2017stress} & SVM & $0.864$ & \cellcolor{gray!25} & \cellcolor{gray!25} & \cellcolor{gray!25} & \cellcolor{gray!25}\\ \hline 
            \cite{verma2019comprehensive} & BBN & $0.926$ & \cellcolor{gray!25} & \cellcolor{gray!25} & \cellcolor{gray!25} & \cellcolor{gray!25} \\ \hline  
            \multirow{2}{*}{\textbf{Our}} & \makecell[l]{CNN w/o self-sup.} & $\textbf{0.984}$ & $0.958$ &  $\textbf{0.957}$ & $0.961$ & $0.956$ \\ \cline{2-7}  
             & \textbf{\makecell[l]{CNN with self-sup.}} & $0.983$ & $\textbf{0.960}$ & $0.956$ & $\textbf{0.963}$ & $\textbf{0.958}$ \\ \hline 
    \end{tabular}
    \label{tab:ssl_swell}
\end{table}

\begin{table}[!t]
    \centering
    \footnotesize
    \caption{The results of our self-supervised method on the AMIGOS dataset are presented and compared to prior work as well as the emotion recognition network without the self-supervised step.}
    \begin{tabular}{l|l|l|l|l|l}
        \hline
            \multirow{2}{*}{\textbf{Ref.}} & \multirow{2}{*}{\textbf{Method}} & \multicolumn{2}{l|}{\textbf{Arousal}} & \multicolumn{2}{l}{\textbf{Valence}} \\ \cline{3-6}
             && \textbf{Acc.} & \textbf{F1} & \textbf{Acc.} & \textbf{F1} \\ \hline\hline
            \cite{amigos_dataset} & GNB & \cellcolor{gray!25}  & $0.545$ & \cellcolor{gray!25}  & $0.551$ \\ \hline
            \cite{santamaria2018using} & CNN & $0.81$  & $0.76$ & $0.71$  & $0.68$ \\ \hline 
            \multirow{2}{*}{\textbf{Ours}} & \makecell[l]{CNN w/o self-sup.} & $0.837$  & $ 0.828$ & $0.809$  & $0.808$ \\ \cline{2-6}
             & \textbf{\makecell[l]{CNN with self-sup.}} & $\textbf{0.858}$  & $\textbf{0.851}$ & $\textbf{0.840}$  & $\textbf{0.837}$ \\ \hline
    \end{tabular}
    \label{tab:ssl_amigos}
\end{table}

\section{Conclusions and Future Work}
We performed emotion recognition using ECG signals with self-supervised learning. First, a network was trained in self-supervised manner to recognize a number of simple signal transformations. The weights from the convolutional layers of this network were then transferred to an emotion recognition network and the fully-connected layers were trained using the original dataset. Our approach achieved state-of-the-art performance on SWELL and AMIGOS datasets and showed that using significantly less data, the self-supervised approach can yield acceptable results. 

For future work, we will further analyze different components of the self-supervised architecture and determine how individual transformation recognition tasks contribute towards the final emotion recognition outcome. Lastly, the contribution of individual network layers to the classification of downstream tasks will be studied.

\begin{small} 
\bibliographystyle{IEEEbib}
\bibliography{strings,refs}

\begin{thebibliography}{10}

\bibitem{picard2001toward}
Rosalind~W. Picard, Elias Vyzas, and Jennifer Healey,
\newblock ``Toward machine emotional intelligence: Analysis of affective
  physiological state,''
\newblock {\em IEEE Transactions on Pattern Analysis \& Machine Intelligence},
  , no. 10, pp. 1175--1191, 2001.

\bibitem{nardelli2015recognizing}
Mimma Nardelli, Gaetano Valenza, Alberto Greco, Antonio Lanata, and
  Enzo~Pasquale Scilingo,
\newblock ``Recognizing emotions induced by affective sounds through heart rate
  variability,''
\newblock {\em IEEE Transactions on Affective Computing}, vol. 6, no. 4, pp.
  385--394, 2015.

\bibitem{pritam_acii}
Pritam Sarkar, Kyle Ross, Aaron~J Ruberto, Dirk Rodenburg, Paul Hungler, and
  Ali Etemad,
\newblock ``Classification of cognitive load and expertise for adaptive
  simulation using deep multitask learning,''
\newblock {\em arXiv preprint arXiv:1908.00385}, 2019.

\bibitem{nussinovitch2011reliability}
Udi Nussinovitch, Keren~Politi Elishkevitz, Keren Katz, Moshe Nussinovitch,
  Shlomo Segev, Benjamin Volovitz, and Naomi Nussinovitch,
\newblock ``Reliability of ultra-short ecg indices for heart rate
  variability,''
\newblock {\em Annals of Noninvasive Electrocardiology}, vol. 16, no. 2, pp.
  117--122, 2011.

\bibitem{healey2005detecting}
Jennifer~A Healey and Rosalind~W Picard,
\newblock ``Detecting stress during real-world driving tasks using
  physiological sensors,''
\newblock {\em IEEE Transactions on Intelligent Transportation Systems}, vol.
  6, no. 2, pp. 156--166, 2005.

\bibitem{liu2009dynamic}
Changchun Liu, Pramila Agrawal, Nilanjan Sarkar, and Shuo Chen,
\newblock ``Dynamic difficulty adjustment in computer games through real-time
  anxiety-based affective feedback,''
\newblock {\em International Journal of Human-Computer Interaction}, vol. 25,
  no. 6, pp. 506--529, 2009.

\bibitem{sensor_kyle}
Kyle Ross, Pritam Sarkar, Dirk Rodenburg, Aaron Ruberto, Paul Hungler, Adam
  Szulewski, Daniel Howes, and Ali Etemad,
\newblock ``Toward dynamically adaptive simulation: Multimodal classification
  of user expertise using wearable devices,''
\newblock {\em Sensors}, vol. 19, no. 19, pp. 4270, 2019.

\bibitem{raina2007self}
Rajat Raina, Alexis Battle, Honglak Lee, Benjamin Packer, and Andrew~Y Ng,
\newblock ``Self-taught learning: transfer learning from unlabeled data,''
\newblock in {\em Proceedings of the 24th International Conference on Machine
  Learning}. ACM, 2007, pp. 759--766.

\bibitem{wang2017transitive}
Xiaolong Wang, Kaiming He, and Abhinav Gupta,
\newblock ``Transitive invariance for self-supervised visual representation
  learning,''
\newblock in {\em Proceedings of the IEEE International Conference on Computer
  Vision}, 2017, pp. 1329--1338.

\bibitem{swell_dataset}
Saskia Koldijk, Maya Sappelli, Suzan Verberne, Mark~A Neerincx, and Wessel
  Kraaij,
\newblock ``The swell knowledge work dataset for stress and user modeling
  research,''
\newblock in {\em Proceedings of the 16th International Conference on
  Multimodal Interaction}. ACM, 2014, pp. 291--298.

\bibitem{amigos_dataset}
Juan~Abdon {Miranda Correa}, Mojtaba~Khomami {Abadi}, Nicu {Sebe}, and Ioannis
  {Patras},
\newblock ``Amigos: A dataset for affect, personality and mood research on
  individuals and groups,''
\newblock {\em IEEE Transactions on Affective Computing}, pp. 1--1, 2018.

\bibitem{kocabas2019self}
Muhammed Kocabas, Salih Karagoz, and Emre Akbas,
\newblock ``Self-supervised learning of 3d human pose using multi-view
  geometry,''
\newblock {\em arXiv preprint arXiv:1903.02330}, 2019.

\bibitem{wu2010open}
Fei Wu and Daniel~S Weld,
\newblock ``Open information extraction using wikipedia,''
\newblock in {\em Proceedings of the 48th Annual Meeting of the Association for
  Computational Linguistics}. Association for Computational Linguistics, 2010,
  pp. 118--127.

\bibitem{tagliasacchi2019self}
Marco Tagliasacchi, Beat Gfeller, F{\'e}lix de~Chaumont Quitry, and Dominik
  Roblek,
\newblock ``Self-supervised audio representation learning for mobile devices,''
\newblock {\em arXiv preprint arXiv:1905.11796}, 2019.

\bibitem{multitask_selfsupervised}
Aaqib Saeed, Tanir Ozcelebi, and Johan Lukkien,
\newblock ``Multi-task self-supervised learning for human activity detection,''
\newblock {\em Proceedings of the ACM on Interactive, Mobile, Wearable and
  Ubiquitous Technologies}, vol. 3, no. 2, pp. 61, 2019.

\bibitem{Xu_2019_CVPR}
Dejing Xu, Jun Xiao, Zhou Zhao, Jian Shao, Di~Xie, and Yueting Zhuang,
\newblock ``Self-supervised spatiotemporal learning via video clip order
  prediction,''
\newblock in {\em The IEEE Conference on Computer Vision and Pattern
  Recognition (CVPR)}, June 2019.

\bibitem{um2017data}
Terry~Taewoong Um, Franz Michael~Josef Pfister, Daniel Pichler, Satoshi Endo,
  Muriel Lang, Sandra Hirche, Urban Fietzek, and Dana Kuli{\'c},
\newblock ``Data augmentation of wearable sensor data for parkinson's disease
  monitoring using convolutional neural networks,''
\newblock {\em arXiv preprint arXiv:1706.00527}, 2017.

\bibitem{TMSi13:online}
``Mobi device (tmsi),'' [Online]. Available:
  \url{http://www.tmsi.com/en/products/mobi.html},
\newblock [Accessed: 2019-09-17].

\bibitem{shimmer}
``Shimmer ecg,'' [Online]. Available:
  \url{http://www.shimmersensing.com/products/shimmer3-ecg-sensor},
\newblock [Accessed: 2019-09-17].

\bibitem{kingma2014adam}
Diederik~P Kingma and Jimmy Ba,
\newblock ``Adam: A method for stochastic optimization,''
\newblock {\em arXiv preprint arXiv:1412.6980}, 2014.

\bibitem{santamaria2018using}
Luz Santamaria-Granados, Mario Munoz-Organero, Gustavo Ramirez-Gonzalez, Enas
  Abdulhay, and NJIA Arunkumar,
\newblock ``Using deep convolutional neural network for emotion detection on a
  physiological signals dataset (amigos),''
\newblock {\em IEEE Access}, vol. 7, pp. 57--67, 2018.

\bibitem{verma2019comprehensive}
Prabal Verma and Sandeep~K Sood,
\newblock ``A comprehensive framework for student stress monitoring in
  fog-cloud iot environment: m-health perspective,''
\newblock {\em Medical \& Biological Engineering \& Computing}, vol. 57, no. 1,
  pp. 231--244, 2019.

\bibitem{sriramprakash2017stress}
S~Sriramprakash, Vadana~D Prasanna, and OV~Ramana Murthy,
\newblock ``Stress detection in working people,''
\newblock {\em Procedia Computer Science}, vol. 115, pp. 359--366, 2017.

\bibitem{koldijk2016detecting}
Saskia Koldijk, Mark~A Neerincx, and Wessel Kraaij,
\newblock ``Detecting work stress in offices by combining unobtrusive
  sensors,''
\newblock {\em IEEE Transactions on Affective Computing}, vol. 9, no. 2, pp.
  227--239, 2016.

\bibitem{alberdi2018using}
Ane Alberdi, Asier Aztiria, Adrian Basarab, and Diane~J Cook,
\newblock ``Using smart offices to predict occupational stress,''
\newblock {\em International Journal of Industrial Ergonomics}, vol. 67, pp.
  13--26, 2018.

\bibitem{harper2019bayesian}
Ross Harper and Joshua Southern,
\newblock ``A bayesian deep learning framework for end-to-end prediction of
  emotion from heartbeat,''
\newblock {\em arXiv preprint arXiv:1902.03043}, 2019.

\end{thebibliography}
\end{small}

\end{document}